\documentclass[runningheads]{llncs}
\usepackage{graphicx}
\usepackage[table]{xcolor}
\usepackage{colortbl}
\usepackage{todonotes}
\usepackage{amsmath}
\usepackage{amssymb}
\usepackage{caption}
\usepackage{subcaption}
\usepackage[toc,page]{appendix}
\usepackage{floatrow}

\usepackage{blindtext}

\usepackage{booktabs}
\usepackage{wrapfig,lipsum,booktabs}

\newcommand{\otoprule}{\midrule[\heavyrulewidth]}
\usepackage[ruled]{algorithm2e}

\usepackage{color}
\usepackage{multirow}


\begin{document}
%

\title{TransforMesh: A Transformer Network for Longitudinal modeling of Anatomical Meshes}
%
%
\author{Ignacio Sarasua, Sebastian P{\"{o}}lsterl,Christian Wachinger,for the Alzheimer's Disease Neuroimaging}
%
\authorrunning{Sarasua et al.}
%
\institute{%
    Artificial Intelligence in Medical Imaging (AI-Med),
    Department of Child and Adolescent Psychiatry
    Ludwig-Maximilians-Universit{\"{a}}t
}

\maketitle              

\begin{abstract}
The longitudinal modeling of neuroanatomical changes related to Alzheimer's disease (AD) is crucial for studying the progression of the disease. To this end, we introduce TransforMesh, a spatio-temporal network based on transformers that  models longitudinal shape changes on 3D anatomical meshes. 
While transformer and mesh networks have recently shown impressive performances in natural language processing and computer vision, their application to medical image analysis has been very limited. 
To the best of our knowledge, this is the first work that combines transformer and mesh networks. 
Our results show that TransforMesh can model shape trajectories better than other baseline architectures that do not capture temporal dependencies. 
Moreover, we also explore the capabilities of TransforMesh in detecting structural anomalies of the hippocampus in patients developing AD.





\end{abstract}

\section{Introduction}

Alzheimer's disease (AD) is a neurodegenerative disease characterized by
progressive cognitive impairment due to neuronal loss (brain atrophy).
To prevent or slow-down cognitive decline, current research suggests
that it is critical to begin treatment before widespread brain damage occurs~\cite{Mehta2017}.
An emerging direction of research focuses on using machine learning to predict 
patient-specific biomarker trajectories to determine a
patient's expected rate of cognitive decline, without relying
on clinical diagnosis (e.g.~\cite{Bilgel2016,Marinescu2019}).
Of particular interest are changes to the hippocampus, because it is among
the first brain structures to show signs of atrophy~\cite{Jack2013}.
If we could reliably predict neuroanatomical changes in the hippocampus,
we could derive a powerful predictor for the expected rate of cognitive decline.

Since temporal modeling requires longitudinal patient data for training,
a major challenge is how to deal with the data heterogeneity:
patients enroll at different stages of the disease, drop-out
at different time points, and might miss a number of intermediate follow-up visits
before returning.
Therefore, each patient's trajectory will differ in terms of length and time
between recorded visits.
To tackle these challenges and satisfy our first requirement of modeling
neuroanatomical changes in the hippocampus, we seek a
method that learns from heterogeneous trajectories and predicts spatio-temporal
changes in the hippocampus. 
In addition, we need to find a shape representation of the hippocampus that is sensitive to small changes and lends itself to deep learning. 

In this paper, we propose the TransforMesh, a longitudinal mesh autoencoder that incorporates
heterogeneous trajectories of varying lengths and missing visits via a transformer model,
and predicts neuroanatomical changes by modifying a mesh representation of the
patient's hippocampus.
To the best of our knowledge, this is the first spatio-temporal model
that uses transformers for longitudinal neuroanatomical shape analysis
that can directly forecast changes to the 3D shape of the hippocampus.
In our experiments, we show that our proposed method is able to predict future trajectories with lower reconstruction error than methods that do not include temporal information. In addition, we have also observed that our model is able to capture very fine changes on shapes belonging to patients developing AD.

\subsubsection{Related work.}
Previous approaches on longitudinal medical image analysis
often use a convolutional neural network (CNN)
to independently extract image descriptors in an encoding step,
which are subsequently passed to a recurrent neural network (RNN)
that processes the sequence to form a latent vector
summarizing the entire image sequence.
This vector can be used as a mean to aggregate the entire
sequence
\cite{campanella2019clinical,cui2019rnn,gao2019distanced,perek2019learning,santeramo2018longitudinal,Xu2019} for the purpose of classifying the latest image in the sequence,
or as input to another RNN to produce a 
sequence of predictions over time \cite{pmlr-v115-hwang20a,yang2017deep}.
All these works use longitudinal image information for classification,
except the work in \cite{pmlr-v115-hwang20a} that uses
a brain connectivity graph as input and produces
another sequence of graphs as output.

While RNNs are used in all works on learning from
longitudinal medical images, they have since been deprecated
in natural language processing (NLP) by the breakthrough of Vaswani~et~al.~\cite{Vaswani2017}
and their Transformer network.
Transformers are now the de-facto standard in NLP, but it has
only been recently that they have been applied to 
natural images~\cite{dosovitskiy2020-vit}.
Their application to medical image analysis is
extremely challenging, because
they require enormous amounts of training data~\cite{Devlin2018-bert,dosovitskiy2020-vit,Lewis2020-bart}.


Prior work on AD classification and shape generation has operated on point clouds~\cite{gutierrez2021discriminative}.
However, point cloud representations can be limited for capturing subtle shape changes and therefore mesh representations have become more popular because they provide an efficient, non-uniform representation of a shape.
Based on these benefits of meshes, several deep neural networks have recently been proposed in computer vision for learning on meshes~\cite{ranjan2018coma,gong2019spiralnet++,hanocka2019meshcnn,feng2019meshnet}. 
Apart from deep learning, the advantage of meshes for identifying anatomical changes over a large population have earlier been noted~\cite{cong2014building}.  



\section{Method \label{sec:method}} 
\begin{figure}[t]
	\centering
	\includegraphics[width=\linewidth]{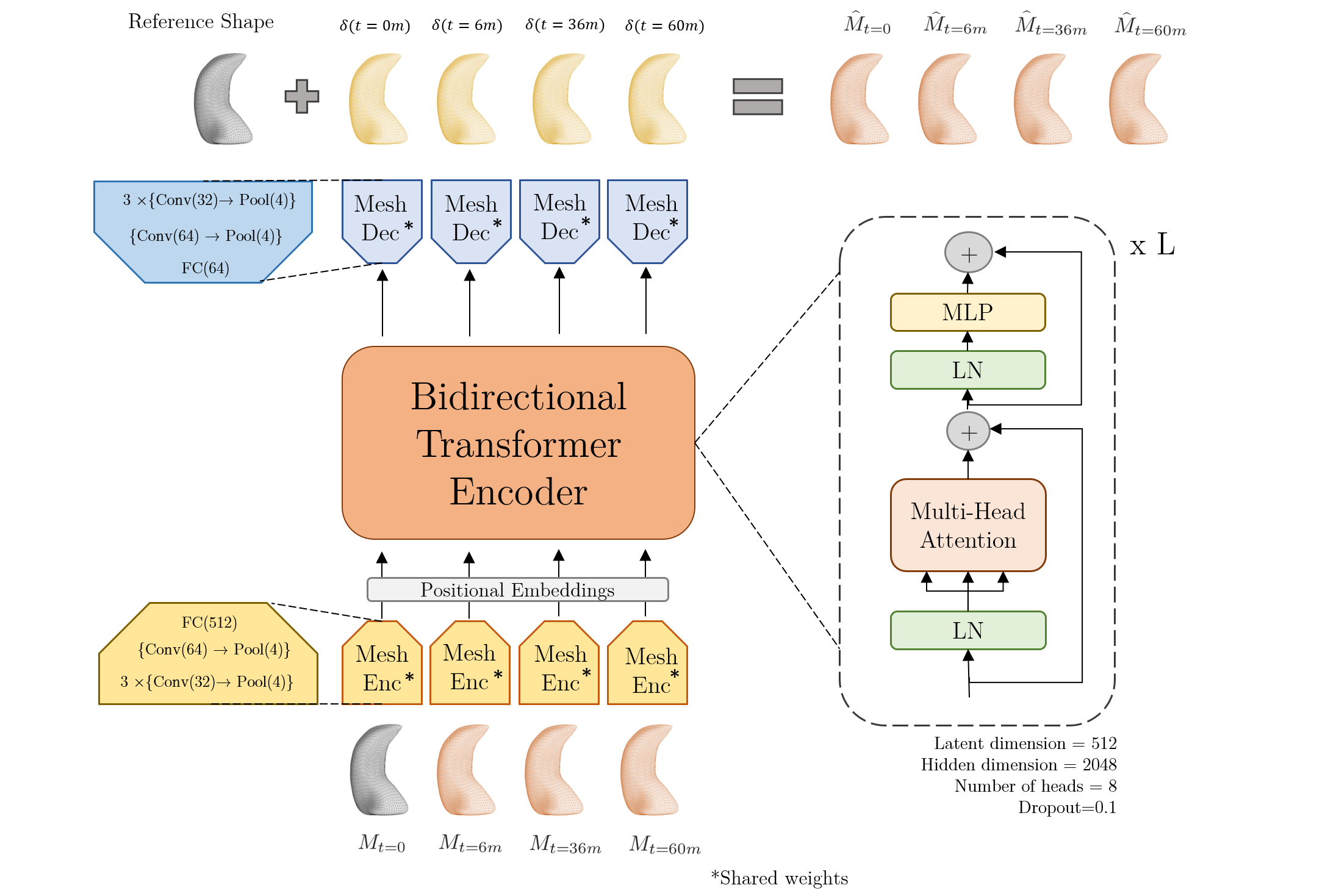}
	\caption{TransforMesh overview: A shared mesh encoder block extracts the latent features from the input sequence of meshes (orange). The Bidirectional Transformer Encoder models the time trajectory and outputs the reconstructed set of latent vectors, which are fed into a shared decoder block that generates the deformations (yellow) that are applied to the reference mesh (gray). The model is trained end-to-end in an autoencoder manner. Details about the architectures are written next to each block. Further information about the parameters can be found in \cite{gong2019spiralnet++} and \cite{Vaswani2017}}
	\label{fig:overview}
\end{figure}

Let $\{M_t^{i} \mid t=0,...,T_i\}$ be the set of hippocampi meshes of subject $i$ from the time of enrollment $t=0$ to the last visit $t=T_i$. One of the biggest challenges of longitudinal studies is that not every subject remains in the study for the same amount of time, nor have all attended every follow-up visit. 
The goal of our method is to model longitudinal shape changes on this highly heterogeneous data. 
An overview of our proposed method can be found in Figure~\ref{fig:overview}. First, the \emph{mesh encoder} extracts latent vectors from each mesh.
Second, the latent vectors  are fed into a \emph{bidirectional transformer encoder}, which models the time dependency between the latent representations. Finally, the \emph{mesh decoder} outputs the reconstructed meshes. The network is trained in an autoencoder manner similar to \cite{Lewis2020-bart}. We provide more details about this training strategy in section~\ref{sec:training}.

\subsection{Mesh Network}\label{sec:mesh_network}
A 3D mesh, $M = (V,E,A,F)$, is defined by a set of vertices, $V\in\mathbb{R}^{N\times 3}$, and edges, $E$, that connect the vertices. $A\in\{0,1\}^{N \times N}$ is the adjacency matrix that indicates the connection between two vertices (e.g. $A_{i,j}=1$ if vertex $i$ is connected to vertex $j$ and $0$ otherwise) and $F$ are the faces formed by a set of edges (three in case of triangular meshes). As opposed to images, meshes are not represented on a regular grid of discrete values and therefore common CNN operations such as \emph{convolution} and \emph{pooling}  are not explicitly defined anymore. In particular, defining local neighborhood for a given vertex becomes challenging.

Gong et al.~\cite{gong2019spiralnet++} proposed SpiralNet++, a novel \emph{message passing} approach to deal with irregular representations, such as meshes and it has achieved state of the art performance in computer vision tasks, such as reconstruction and classification. To the best of our knowledge, this work is the first application of SpiralNet++ for anatomical shape analysis. SpiralNet++ encoder and decoder blocks are formed by three main operations: spiral convolution, mesh pooling, and un-pooling.

\begin{figure}[t]
    \centering

    \includegraphics[width=\textwidth]{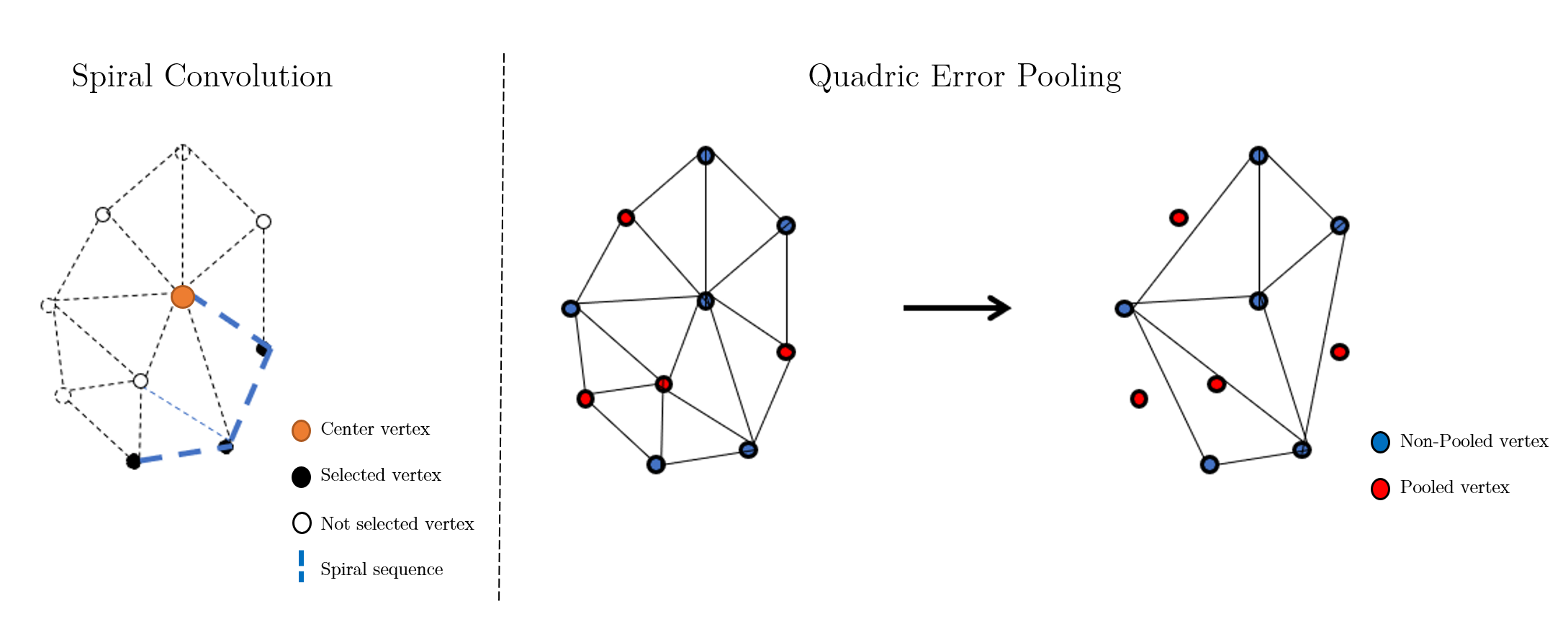}
    \caption{Left: Definition of a spiral sequence given a vertex $v$ (orange); Right: Quadric Pooling operation used by SpiralNet++. Mesh features are down-sampled by removing red vertices that minimize quadric error.}

    \label{fig:conv_pooling}
\end{figure}

\hfill\\
\underline{Spiral convolution:}
Due to the nature of triangular meshes, a spiral serialization of neighboring nodes is possible. Given a vertex  in $V$, \cite{gong2019spiralnet++} defines its spiral sequence by choosing an arbitrary starting direction in counter-clockwise manner. An example of a spiral sequence can be found in Figure~\ref{fig:conv_pooling}. In comparison to SpiralNet
\cite{lim2018simple}, SpiralNet++ defines these sequences only once for a template shape and then applies them to the aligned samples in the dataset, highly increasing the efficiency of the method. 
The convolution operation in layer $k$ for features $\mathbf{x}_i$ associated to the $i$-th vertex is therefore defined as:
$
    \mathbf{x}_i^{(k)} = \gamma^{(k)}\left(\underset{j \in S(i,l)}{\parallel} \mathbf{x}_j^{(k-1)} \right),
$
where $\gamma$ denotes MLPs and $\parallel$ is the concatenation operation. $S(i,l)$ is an
ordered set consisting of $l$ vertices inside the spiral. More details about the method and its implementation can be found in \cite{gong2019spiralnet++}.


\hfill\\
\underline{Mesh Pooling and Un-pooling}: The \emph{pooling} is obtained by iteratively contracting vertex pairs that would minimize the quadratic error \cite{garland1997surface}. In Figure \ref{fig:conv_pooling}, we illustrate this process. For efficiency, the coordinates of the vertices that must be pooled in each level are computed for the template and then applied to the samples in the dataset. The un-pooling operation is done by using the barycentric coordinates of the closest triangle in the downsampled mesh \cite{ranjan2018generating}.

\subsection{Transformer Network}\label{sec:rec_blocks}
Figure~\ref{fig:overview} shows the transformer encoder network, which is inspired by the one proposed in \cite{dosovitskiy2020-vit}. The Bidirectional Transformer Encoder is formed by concatenating $L$ encoder blocks. Each encoder block is formed by a Multi-Head-Attention and an MLP, each preceded by a Layer Normalization (LN) block. In addition, residual connections are added after every block and GELU is used as non-linearity in the MLP block. Following \cite{dosovitskiy2020-vit}, we also use learnable 1D position embeddings to retain positional information. 

\subsection{Training procedure \label{sec:training}}
As illustrated in Figure \ref{fig:overview}, our model predicts the mesh deformation field $\delta(t)$ with respect to a \emph{reference mesh}, which is a person's baseline scan $M_{t=0}^i$.

\textbf{Missing shapes: } In order to account for variable lengths within the sequences, transformer networks use padding tokens.
These are ignored during training by adding key masks and disregarded when computing the regression loss. For our application, we use the encoding of the reference shape, as produced by the mesh network, to generate the ``missing value'' embedding.

\textbf{Data Augmentation: } Since the goal of our method is to model longitudinal shape change in the presence of missing values, some shapes are randomly selected and substituted by the reference shape. Contrary to the real missing shapes, these are not ignored during training: the self-attention masks do take them into consideration and they are included when computing the final regression loss.

\textbf{Loss function: } To account for the drop in the number of scans as the study goes on, we introduce an exponential factor that gives more weight to scans  further from the reference. The training objective function is defined as:
$
    \mathcal{L} = \sum_{i=0}^{P} \sum_{t=0}^{T_i}  e^t||M_{t=0} + \delta(t) - M_{t}|| - \alpha||\delta(t)||,    
$
where $\alpha$ (empirically set to $10^{-4}$) is a regularization constant that prevents the network from copying the reference shape, and $P$ is the number of patients in the set.


\section{Experiments}
For all our experiments, we use the shape of the left hippocampus, given its  importance in AD pathology~\cite{Jack2013}. We use data from the Alzheimer’s Disease Neuroimaging Initiative (ADNI) database (adni.loni.usc.edu)~\cite{Jack2008}.
Structural scans are segmented with FIRST~\cite{Patenaude2011} from  the FSL Software Library, which provides  meshes for the segmented samples. 
FIRST segments the subcortical structures by registering them to a reference template, creating voxel-wise correspondences (and therefore, also vertex-wise) between the template and every sample in the dataset. We use this template as the reference for all the pre-computations of the mesh network in Sec.~\ref{sec:mesh_network}. We limit the number of follow-ups $S = 8$ (from baseline to 72 months) and the latent dimension for the Mesh Encoder $D=512$. Our complete dataset consists of  1247 patients split 70/15/15 (train/validation/test) following a data stratification strategy that accounts for age,sex and diagnosis, so they are represented in the same proportion in each set. The average number of follow-up scans per patient is $3.25$ and only a 3\% of the patients have attended all the follow-up sessions. 

\subsection{Implementation details}
We define 5 network architectures to evaluate our experiments. First, we define three versions of our method: Tiny TransforMesh (TTM), Small TransforMesh (STM) and  Base TransforMesh (BTM) with depths L=1, L=3 and L=12, respectively. Further, we define a baseline and an upper-bound method.

\textbf{Baseline} 
The main contribution of transformer networks is their capability of modeling the time dimension in the latent space. To measure the effect of this, we design a baseline method that does not take the time component into account. Inspired by the work in \cite{gutierrez2021discriminative}, after the mesh encoder path,  the transformer network is substituted by a fully connected bottle neck (FCBN) formed by 3 MLP blocks (dimensions are $[S \times D, S ,S \times D]$ respectively). The latent vectors obtained by the mesh encoder are concatenated and passed to the FCBN. The output is decoded as in our method. Notice that, as in \cite{gutierrez2021discriminative}, this method does not capture the temporal component (only the spatial). Contrary to our method, this method is not flexible to any sequence length ($S$), since the FCBN dimensions directly depend on this parameter.

\textbf{Mesh Autoencoder}
In order to put the reconstruction error into context, we train a mesh autoencoder, MeshAE, to reconstruct the shapes, which uses the same encoder and decoder architecture as our model.  
The main difference between MeshAE and our model is that MeshAE is only reconstructing the shape from its latent representation, while our model has never seen the shape. 
Since the mesh part of both networks are identical, we expect our network to be limited by this method. Details about the architecture can be found in Figure \ref{fig:overview}.



\subsection{Longitudinal shape modeling}
Given the set of meshes belonging to a subject, $\{M_t^{i} \mid t=0,...,T_i\}$, the goal of our method is to predict missing shapes. We divide our experiments in three scenarios:  shape interpolation,  shape extrapolation and trajectory prediction.

\textbf{Interpolation:} In this first experiment, we aim to interpolate the mesh of an intermediate follow-up session, giving future and past shapes. From every patient in the test set, we remove from the sequence the shape in the middle $M_{t=\mu}^i$ with $\mu = \lfloor T_i/2 \rfloor$ and pass the sequence through the network. We compute the interpolation error as the Mean Absolute Error (MAE) between $M_{t=\mu}^i$ and $\hat{M}_{t=\mu}^i$, where  $\hat{M}_{t=\mu}^i$ is the predicted mesh for $t=\mu$. 

\textbf{Extrapolation:} In the second experiment, the goal is to predict the mesh of the last available shape, based on all the previous ones. Hence, from every patient in the test set, we remove from the sequence the shape $M_{t=T_i}^i$ and input the remaining shapes to the network. The extrapolation MAE is computed between $M_{t=T_i}^i$ and $\hat{M}_{t=T_i}^i$.  

\textbf{Trajectory prediction:} The third experiment is similar to the extrapolation experiment, but it predicts shapes that are more distant in time. Therefore, we only input into the network the shape belonging to the baseline scan $M_{t=0}$ and predict all the shapes that are at least 2 years apart. We define the future error as $FME = \mathbb{E} \left[ \parallel \hat{M}_{t} - M_{t} \parallel |24m \leq t \leq T_i\right]$.


\subsection{Results}
Table \ref{table:results} reports the median error and the median absolute deviation along all the patients and scans.
We observe that all the Transformer models bring a significant improvement with respect to the baseline method, confirming our initial hypothesis regarding the importance of spatio-temporal models. It is also worth mentioning that the most shallow transformer network (TTM) is able to predict the missing shapes better than the baseline method while having 20\% less trainable parameters.

For the extrapolation and trajectory experiments, MeshAE yielded better results, as expected. However, for the interpolation experiment, the temporal methods yielded a lower median error than MeshAE. These results indicate that it could be advantageous to have access to the longitudinal sequence even if the mesh from the actual time point T is not included, especially when T is close to 0. Yet, the increased deviation also indicates that the variation of the temporal models is higher.

\begin{table}[t]
\centering
\caption{The table reports the median error and median absolute deviation for the interpolation, extrapolation, and trajectory experiments. Errors were multiplied by $10^2$ to facilitate presentation. We compare to the mesh autoencoder (MeshAE), the fully connected bottle neck (FCBN), and tiny, small, and base TransforMesh. Also reported are the model's number of parameters. }
\begin{tabular}{l@{\hskip 0.09in}c@{\hskip 0.07in}c@{\hskip 0.07in}c@{\hskip 0.07in}r}

\toprule
Method            & Interpolation                       & Extrapolation                      & Trajectory                       & \#params  \\
            \otoprule
MeshAE     & $22.2  \pm 1.5$          & $22.4 \pm 1.7$          & $22.2 \pm 1.7$                 & $0.9M$            \\
\midrule
FCBN & $20.4 \pm 6.0$ & $26.0 \pm 7.7$          & $28.0 \pm 8.9$  & $5.1M$       \\
\midrule
TTM (1 block)  & $20.2 \pm 5.5$ & $25.7 \pm 7.5$ & $27.7\pm 8.9$ & $4.1M$     \\
STM (3 blocks)& $20.3 \pm 5.6$ & $25.5 \pm 7.3$          & $27.6 \pm 8.9$  & $10.4M$    \\
BTM (12 blocks) & $20.3 \pm 5.6$ & $ 25.4 \pm 7.4$          & $ 27.7 \pm 8.9$  & $38.7M$   \\ 	\bottomrule    
\end{tabular}
\setlength{\belowcaptionskip}{-10pt}
\label{table:results}
\end{table}

\subsection{Anomaly Visualization}
A common application of Autoencoder Networks in medical image analysis is  anomaly detection \cite{baur2021autoencoders}. The idea is training the network on healthy controls and passing patients with anomalies during inference time. 
Given an autoencoders' denoising behaviour, these anomalies can be detected in an unsupervised manner by computing the differences between the original and the reconstructed shapes. 
Those areas with higher reconstruction error are considered anomalies, since they are not part of the healthy training distribution. 

\begin{figure}[t]
    \centering

    \caption{Vertex-wise Absolute Error between the meshes of a patient converting from MCI to dementia and our method's prediction when trained on HC and MCI subjects.}
    \includegraphics[width=\textwidth]{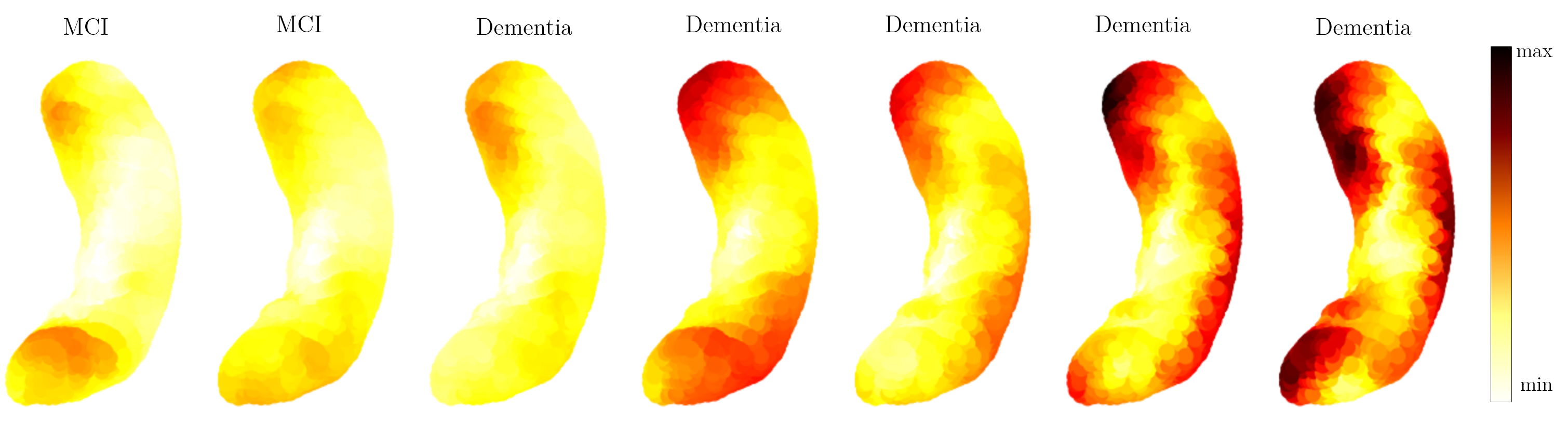}

    \label{fig:hipp_evol}
\end{figure}

We replicate this experiment in our longitudinal setting. To the best of our knowledge, this is the first work exploring longitudinal anomaly detection on anatomical shapes. 
We train our model on subjects that have not been diagnosed with dementia. 
During inference, we include AD patients and compute the error between the input and the output. 
In Figure~\ref{fig:hipp_evol}, we can observe an example belonging to a patient converting from MCI to AD. Even though we do not have an anomaly mask to compare to (like in other techniques \cite{baur2021autoencoders}) we can observe a strong correlation between the areas with highest error (medial part of the body in the subiculumarea,  the lateral part of the body in the CA1 area and the inferior part of the hippocampus head in the subiculum area) and those proven to be more affected my Alzheimer's disease \cite{lindberg2012shape}. 

\section{Conclusion}
In this work, we have proposed the first deep spatio-temporal  model for longitudinal analysis of anatomical shapes. To the best of our knowledge, this is the first approach that combines mesh and transformer networks. Our experiments demonstrated that TransforMesh outperforms the baseline methods, which rely only on spatial features, even on its lightest implementation. This makes it ideal for medical applications where the amount of data is limited. 
We further illustrated that TransforMesh  can be used to detect structural anomalies on patients converting from MCI to AD.

\subsubsection{Acknowledgements}
This research was supported by the Bavarian State Ministry of Science and the Arts and coordinated by the Bavarian Research Institute for Digital Transformation,
and the Federal Ministry of Education and Research in the call for Computational Life Sciences (DeepMentia, 031L0200A).

\bibliographystyle{splncs03.bst}
\bibliography{biblio.bib}
\end{document}